%% file: cameraready.tex
\documentclass[runningheads]{llncs}

\usepackage{eccv}

\usepackage{eccvabbrv}

\usepackage{graphicx}
\usepackage{booktabs}
\usepackage{colortbl}
\definecolor{grey1}{gray}{0.9}
\usepackage{multirow}
\usepackage{caption}
\usepackage[accsupp]{axessibility}  

\usepackage{hyperref}

\usepackage{orcidlink}
\newcommand\samethanks[1][\value{footnote}]{\footnotemark[#1]}

\begin{document}

\title{Distill Once, Adapt Life-Long: Exploring Dataset Distillation for Continual Test-Time Adaptation} 

\titlerunning{DO-ALL}

\author{Hyun-Kurl Jang\inst{1}\orcidlink{0009-0003-7943-3326}\thanks{Equal contribution.} \and
Jihun Kim\inst{1}\orcidlink{0009-0007-8764-195X}\samethanks \and
Hyeokjun Kweon\inst{2}\orcidlink{0000-0003-4442-5513}\samethanks \and \\ Kuk-Jin Yoon\inst{1}\orcidlink{0000-0002-1634-2756}}

\authorrunning{Jang et al.}

\institute{KAIST, Visual Intelligence Lab \\
\email{\{jhg0001, jihun1998, kjyoon\}@kaist.ac.kr}\\
\and
Chung-Ang University, FoVLab\\
\email{hyeokunkweon@cau.ac.kr}
}

\maketitle

\begin{abstract}
  Continual Test-Time Adaptation (CTTA) aims to maintain model performance under evolving target domains by adapting online without labeled data. 
However, practical deployments often cannot retain the source dataset due to privacy or licensing constraints, and purely source-free CTTA methods tend to become unstable under long-term distribution shift, suffering from compounding self-training errors and catastrophic forgetting.
We introduce \textbf{DO-ALL} (Distill Once, Adapt Life-Long), a plug-and-play framework that revisits source information in a compact and privacy-conscious form via Dataset Distillation (DD). 
Before deployment, DO-ALL performs DD to produce a small set of synthetic distilled anchors that summarize the source distribution. 
During adaptation, each target sample is matched with its most semantically aligned anchor, which provides a stable reference for various CTTA via source replay, representation alignment, and manifold-smoothing regularization.
DO-ALL can be seamlessly integrated into existing CTTA algorithms, consistently improving long-term robustness across CIFAR100-C, ImageNet-C, and the CCC benchmark. 
This demonstrates the potential of leveraging DD to enable stable and continuous adaptation without retaining raw source data.
The code is available at \url{https://github.com/blue-531/DOALL}.
\end{abstract}

\section{Introduction}\label{sec:intro}

Perception systems rarely operate in a static environment. 
After deployment, external factors such as weather can vary, sensors may degrade, and operational contexts often shift. 
Test-Time Adaptation (TTA)~\cite{wang2020tent, li2016revisiting, lim2023ttn, liang2020we} addresses this by updating a source-trained model with unlabeled target data encountered during inference. 
Further, Continual Test-Time Adaptation (CTTA)~\cite{wang2022continual,yuan2023robust,song2023ecotta,marsden2024universal,niu2022efficient,lee2024becotta,dobler2023robust,boudiaf2022parameter,niu2023towards,zhu2024reshaping,liu2024continual,yang2024versatile} considers a more dynamic scenario in which the model must adapt online to a sequence of evolving domains while maintaining prediction quality.

Since distributing or retaining the full source set is often infeasible (\textit{e.g.}, due to privacy/license issues), various CTTA works adopt a source-free setting.
While this direction has clear practical advantages, fully discarding source information fundamentally limits stability under long-term, non-stationary target streams.
As distributions drift, self-training signals become increasingly unreliable, and repeated updates exacerbate catastrophic forgetting.
In such cases, relying solely on the test stream makes long-term stabilization inherently challenging, as the model lacks a persistent anchor to keep its representation grounded~\cite{choi2022improving,dobler2023robust}.

\begin{figure}[t]
    \centering
    \includegraphics[width=0.99\linewidth]{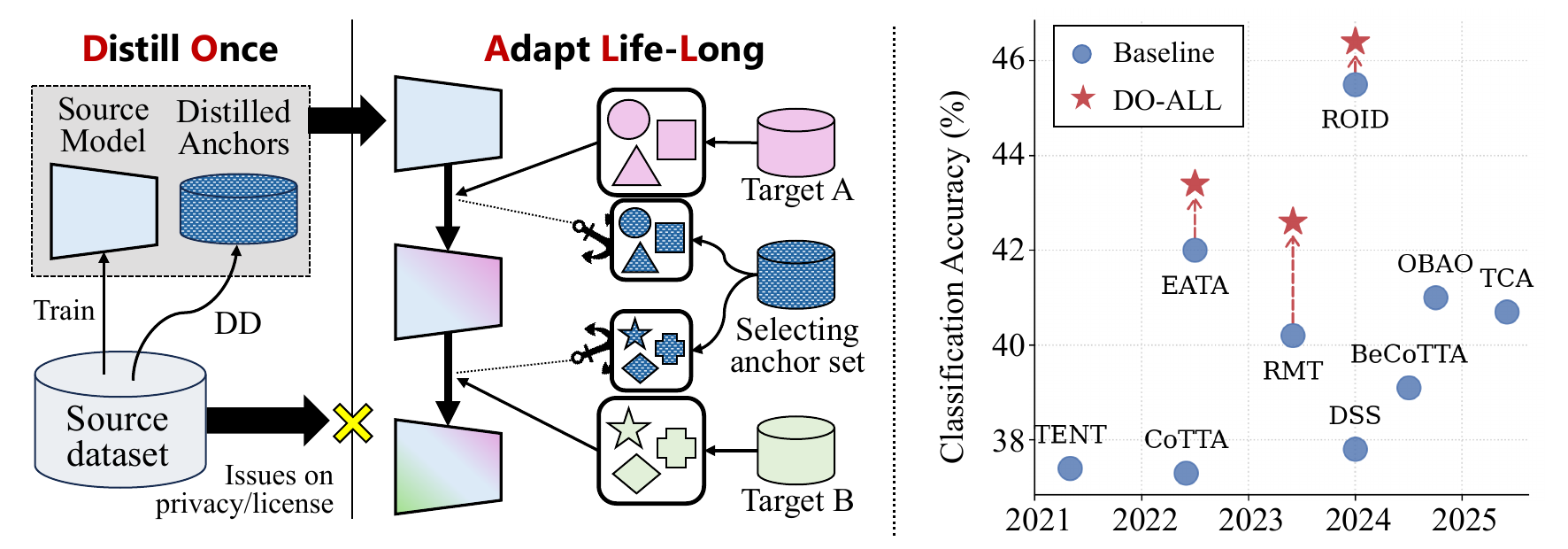}
    \vspace{-5pt}
    \caption{\textbf{Left:} Overview of our \textbf{DO-ALL} framework. Before deployment, we distill the source dataset into a compact set of synthetic anchors. After deployment, each target sample is matched to the anchor that best aligns with its representation. The selected anchors serve as stable reference points that regularize model updates and prevent drift throughout CTTA. \textbf{Right:} Chronological comparison of accuracy on ImageNet-C for existing CTTA methods (blue circles) and DO-ALL-augmented variants (red stars).}
    \vspace{-15pt}
    \label{fig:intro}
\end{figure}

Several works~\cite{dobler2023robust, kang2023leveraging,adachi2023covariance,jung2023cafa,adachi2024test,su2024unraveling} have explored using source information in ways that avoid distributing the raw dataset.
For example, they have proposed to retain only source statistics~\cite{jung2023cafa,lim2023ttn}, feature prototypes~\cite{dobler2023robust,choi2022improving}, or lightweight proxy representations~\cite{kang2023leveraging,dobler2023robust,adachi2024test}. 
These strategies aim to provide a stable reference that prevents the model from drifting excessively during adaptation, while still respecting the practical constraints of the source-free setting. 
In a similar spirit, we revisit source information in a compact and privacy-conscious manner. 

Our \textbf{DO-ALL} (Distill Once, Adapt Life-Long) framework is shown in Fig.~\ref{fig:intro}. 
It performs Dataset Distillation (DD) on the source dataset once before deployment, summarizing the source domain into a tiny set of synthetic samples.
DD is known to compress a large training set into a small number of learnable synthetic examples that preserve class-discriminative structure and training behavior, while requiring orders of magnitude less storage and exposing far less raw data content~\cite{wang2018dataset,zhao2021dataset}.
Thus, the distilled samples provide a compact and persistent representation of the source distribution that can be carried into deployment.

A key advantage of DO-ALL is that it is designed to integrate this distilled source knowledge into existing CTTA algorithms without altering their objectives or architectures. 
Therefore, the distilled samples function as an \textbf{optional} stability buffer that can be plugged into a broad range of CTTA methods to counteract drift and forgetting, rather than replacing their adaptation strategy.

To achieve this, we introduce two core components in DO-ALL.
First, before deployment, we construct source-distilled anchors via DD (left of Fig.~\ref{fig:intro}).
Each anchor consists of (1) a synthetic sample, (2) its soft label under the source model, and (3) its latent feature in the source feature space. 
These anchors provide compact but semantically meaningful reference points for the model to preserve the source representation, even after deployment.

Second, during CTTA, we associate each target sample with its most semantically related distilled anchor (right of Fig.~\ref{fig:intro}).
This correspondence then guides adaptation through source replay, MixUp-style regularization~\cite{zhang2017mixup}, and feature-space alignment that preserves local neighborhood structure.
To further prevent long-term drift, we introduce a harm-adaptive blending mechanism that selectively restores unstable parameter groups toward the source initialization based on their gradient magnitudes.

Together, these enable \textbf{DO-ALL} to deliver stable and robust adaptation, while preserving \textbf{plug-and-play} compatibility with existing CTTA methods.
We experimentally verify that DO-ALL consistently improves a wide range of CTTA methods when paired with various DD strategies, yielding notable gains across CIFAR100-C~\cite{hendrycks2019benchmarking}, ImageNet-C~\cite{hendrycks2019benchmarking}, and the CCC benchmark~\cite{press2023rdumb}.

Beyond the main results, we conduct extensive ablations on the proposed components, anchor association, and cross-architecture settings.
Interestingly, we observe a clear positive correlation between the quality of the distilled anchors and the magnitude of CTTA improvement in DO-ALL: as DD methods produce more informative synthetic samples, DO-ALL can translate this semantic richness directly into stronger stabilization during adaptation.
We hope this work encourages further exploration at the intersection of DD and CTTA toward more robust deployment-time learning.

\section{Related work}
\label{sec:related_works}
\subsection{Dataset Distillation (DD)}
Scaling model capacity has intensified the demand for massive datasets, which in turn raises storage and compute costs. DD~\cite{cui2022dc,li2025dd,zhao2021dataset,nguyen2020dataset,zhou2022dataset, jeong2026multimodal} tackles this by synthesizing a compact set of images whose training effect approximates that of the full dataset, but early approaches~\cite{wang2018dataset,zhao2021dataset} incurred substantial computation and slow convergence. 
Later methods improved scalability and stability via trajectory matching~\cite{cazenavette2022dataset, zhao2020dataset,lee2022dataset,du2023minimizing,du2023sequential,guo2023towards,liu2024dataset,lee2024selmatch,yang2024neural,zhong2025towards,shin2023loss
}, aligning optimization paths between real and synthetic training to reduce parameter-update mismatch and accelerate convergence.
Distribution matching approaches~\cite{zhang2024dance,zhao2023improved,wang2022cafe,zhao2023dataset,wang2025dataset,liu2023dataset,shen2025delt,sajedi2023datadam, deng2024exploiting,cui2025optical,li2025diversity, zhang2024m3d, li2025hyperbolic} minimize cross-layer feature discrepancies to improve generalization.
These advances have expanded DD to large-scale settings~\cite{zhang2025infer,ma2025curriculum,shen2025delt,yin2023squeeze,liu2023dataset,sun2024information,shao2024elucidating,cui2023scaling} with stronger memory/time profiles. Beyond compression, DD has been adopted in continual learning~\cite{yang2023efficient} and privacy-sensitive regimes~\cite{dong2022privacy, song2023federated, xiong2023feddm} where raw data cannot be shared. Our approach leverages a distilled source set as a compact, privacy-aware anchor within continual test-time adaptation, using it as an explicit anti-forgetting prior that stabilizes adaptation over evolving target distributions.

\subsection{Continual Test-Time Adaptation (CTTA)}

Source-trained models typically degrade under distribution shift at test time. TTA~\cite{wang2020tent,li2016revisiting,lim2023ttn,liang2020we, jang2024talos, kim2025dc, kim2026bootstrapping,kim2026test} addresses this by updating the trained model on unlabeled target samples without accessing source data.
Common strategies include entropy minimization~\cite{wang2020tent}, consistency across stochastic augmentations~\cite{zhang2022memo}, and normalization re-estimation to align activation statistics to the target data~\cite{li2016revisiting,lim2023ttn}.
In practice, however, deployments rarely face a static target; the test distribution often drifts over time. CTTA~\cite{wang2022continual,yuan2023robust,song2023ecotta,marsden2024universal,niu2022efficient,lee2024becotta,dobler2023robust,boudiaf2022parameter,niu2023towards,zhu2024reshaping,liu2024continual,yang2024versatile} formalizes this setting, aiming to adapt over long horizons while avoiding error accumulation and forgetting. 
CoTTA~\cite{wang2022continual} employs a student–teacher framework and stochastic parameter restoration to prevent error accumulation over long streams. RMT~\cite{dobler2023robust} reinforces teacher consistency with contrastive objectives, keeping features close to the source space. EcoTTA~\cite{song2023ecotta} focuses on low-memory updates and self-distilled regularization to keep adaptation lightweight and stable. 
CTTA methods primarily adapt online from unlabeled target streams, but some explicitly carry over source information using cached prototypes, source proxy or feature statistics~\cite{jung2023cafa,lim2023ttn,choi2022improving,kang2023leveraging,dobler2023robust,adachi2024test}. These priors are prepared before test time and then combined with standard TTA objectives  to improve stability. We take this idea a step further by using a distilled source dataset as the carried-over structure: it preserves class-discriminative information in a compact form, enabling regularization when needed.

\section{Method}
In this paper, we explore the potential of DD within the context of CTTA.
Following the conventional formulation of CTTA~\cite{yuan2023robust,marsden2024universal,ni2025maintaining}, let $D_{s} \sim P_{s}$ denotes the source dataset and target inputs are drawn from time-varying distributions $P_t$.
Here, no target labels are available at test time, and the stream is observed only once.
The goal of CTTA is to adapt a source-trained model $f_{\theta_0}$, parameterized by $\theta_0$, to the non-stationary target stream.

\begin{figure*}[t]
    \centering
    \includegraphics[width=0.99\linewidth]{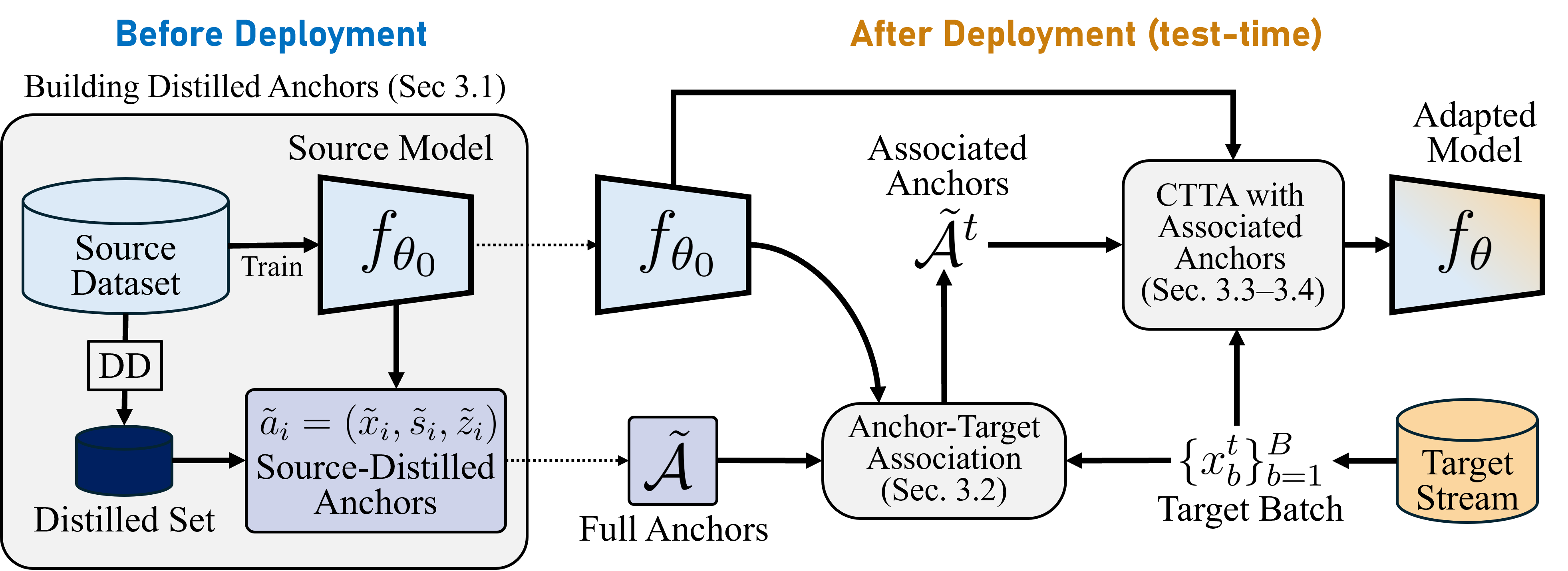}
    \vspace{-5pt}
    \caption{Before deployment, we perform DD once on the source dataset to obtain a compact set of source-distilled anchors $\tilde{\mathcal{A}}$, where each anchor stores a synthetic input, its soft label under the source model, and its latent feature representation (Sec.~3.1).
    After deployment, during CTTA, each incoming target sample is matched to its most semantically aligned anchor in feature space, forming the associated anchor set $\tilde{\mathcal{A}}^t$ for the current batch (Sec.~3.2).
The model is then updated with anchor-guided objectives and stabilized via harm-adaptive blending (Sec.~3.3–3.4).
Together, these components enable stable and robust CTTA with the distilled summary of the source dataset.}
    \vspace{-10pt}
    \label{fig:doall}
\end{figure*}

A major challenge in CTTA is to adapt to continuously shifting target distributions while preserving the knowledge acquired from the source domain, \textit{i.e.}, avoiding catastrophic forgetting~\cite{wang2022continual, press2023rdumb}.
The key idea behind our DO-ALL is to leverage dataset distillation to compactly preserve essential source information, allowing it to serve as a stable \textbf{anchor} during adaptation.
We illustrate the overall framework in Fig.~\ref{fig:doall}, describing (1) what information is distilled and stored from the source domain, and (2) how the stored distilled knowledge is effectively utilized to regularize adaptation throughout the evolving target stream.

\subsection{Building Source-Distilled Anchors through DD}

Dataset distillation (DD)~\cite{wang2018dataset,cazenavette2022dataset,liu2023dataset,yin2023squeeze,shen2025delt} compresses a large dataset into a small set of informative synthetic samples that preserve the training dynamics of the original data.
We apply DD to the source dataset $D_s$ once before deployment, obtaining a compact synthetic set $\tilde{D}_s=\{\tilde{x}_i\}_{i=1}^{N_a}$ with $N_a \ll |D_s|$.
This provides a memory- and privacy-efficient summary of the source distribution during CTTA.
Throughout the paper, we use the tilde notation (\textit{e.g.}, $\tilde{x}$) to denote quantities derived from the distilled source set.

Each distilled sample $\tilde{x}_i$ is assigned a soft pseudo-label by source model:
\begin{equation}\label{eq:soft_label}
\tilde{s}_i = f_{\theta_0}(\tilde{x}_i).
\end{equation}
We use soft rather than hard labels because they preserve the inter-class relations between distilled samples and the class-wise confidence structure encoded by the source model, offering richer semantic guidance for more robust CTTA.

In addition to the soft label, we extract and store the latent representation of each synthetic sample.
We decompose the source model into $f_{\theta_0} = h_{\theta_0} \circ g_{\theta_0}$, where $g_{\theta_0}$ is the feature extractor and $h_{\theta_0}$ the classifier head.
The latent representation of each distilled sample is:
\begin{equation}\label{eq:latent_anchor}
\tilde{z}_i = g_{\theta_0}(\tilde{x}_i).
\end{equation}
These latent features $\{\tilde{z}_i\}$ capture the source feature-space geometry and serve as stable reference points that preserve representational structure during CTTA.

Finally, we package the synthetic input, its soft label, and its latent feature into a single unit, which we refer to as a source-distilled anchor: 
\begin{equation}
 \tilde{a}_i = (\tilde{x}_i, \tilde{s}_i, \tilde{z}_i),
\end{equation}
and denote the full anchor set as $\tilde{\mathcal{A}}=\{\,\tilde{a}_i\,\}_{i=1}^{N_a}$.

This anchor set $\tilde{\mathcal{A}}$ is constructed once before deployment and carried with the model throughout CTTA, providing a persistent reference to the source distribution during adaptation.
Compared to storing the full source dataset, $\tilde{\mathcal{A}}$ is dramatically more compact and privacy-preserving, yet remains sufficiently informative to serve as an effective guide.

\subsection{Anchor-Target Association at Test Time}

After constructing and storing the anchors, we deploy the model and perform continual adaptation on the incoming target stream.
As the target distribution drifts gradually, each test-time batch typically reflects only a local region of the shift.
Therefore, rather than referencing all anchors in $\tilde{\mathcal{A}}$ uniformly, we retrieve the most relevant source information for each target sample based on feature similarity, enabling context-aware and stable adaptation.

Formally, at test time $t$, a mini-batch of $B$ target samples $\{x^t_b\}_{b=1}^{B}$ arrives.
For each target sample $x^t_b$, we compute its latent representation $z^t_b = g_{\theta}(x^t_b)$ and compare it with the anchor features \(\{\tilde{z}_i\}_{i=1}^{N_a}\) using cosine similarity:
\begin{equation}\label{eq:selection}
\text{sim}(b,i) = \frac{z^t_b \cdot \tilde{z}_i}{\|z^t_b\|_2\,\|\tilde{z}_i\|_2}.
\end{equation}
We then assign to each target sample its nearest anchor:
\begin{equation}\label{eq:selection_idx}
j = \arg\max_i \text{sim}(b,i),
\quad
\tilde{a}^t_b = (\tilde{x}_j, \tilde{s}_j, \tilde{z}_j).
\end{equation}

To put it simply, each target sample $x^t_b$ is paired with its nearest anchor $\tilde{a}^t_b$.
Collecting these assignments over the batch yields the associated anchor set $\tilde{\mathcal{A}}^t = \{\,\tilde{a}^t_b\,\}_{b=1}^{B}$ for CTTA at this timestep $t$.
This provides each target sample with a semantically aligned source reference.
By grounding adaptation updates in latent-space similarity, the model adapts within semantically consistent regions of the representation space, substantially improving stability under continual distribution shift.

\subsection{Test-Time Adaptation with Associated Anchors}

Given the associated anchor set $\tilde{\mathcal{A}}^t$ for the current test-time batch, we adapt the model using two complementary objectives.
These respectively (1) preserve source knowledge while imposing local smoothness around the anchor–target manifold and (2) align the evolving representation with the source feature geometry.
All objectives operate on the paired correspondences between target samples and their matched anchors.


\noindent\textbf{(1) Anchor-based replay loss.}
To prevent catastrophic forgetting during continual adaptation, we replay source knowledge through the source-distilled anchors and constrain the model to remain consistent with them. 
Compared to relying on the full source dataset, distilled anchors require no access to \(D_s\) at test time, making them suitable for deployment settings with memory or privacy constraints.
Moreover, unlike randomly sampled real images or coreset-style subsets, each distilled synthetic sample is optimized to serve as a compact and information-dense proxy of the source distribution~\cite{lei2023comprehensive,cui2022dc,li2025dd}, capturing class-level structure more effectively than uniform subsampling or clustering-based selection.
As a result, anchor replay provides stronger and more targeted guidance for stabilizing the model under distribution shift.

Our replay objective consists of two terms.
The first term performs \emph{direct replay} on anchors, preserving the source model's predictive behavior.
The second term extends replay to the \emph{local target--anchor manifold} by mixing each target sample with its matched anchor, which encourages smoother predictions in the region where adaptation is actually taking place.

\emph{(i) Direct replay.}
For each matched anchor \(\tilde{x}_b\) with stored soft label \(\tilde{s}_b\), we enforce consistency via temperature-scaled KL:
\begin{equation}\label{eq:replay_direct}
\mathcal{L}_{\text{dir}}
=\beta_s^2 \cdot \mathrm{KL}\!\left(
\mathrm{softmax}\!\big(f_{\theta}(\tilde{x}_b)/\beta_s\big)\,
\middle\|\,
\mathrm{softmax}\!\big(\tilde{s}_b/\beta_s\big)
\right),
\end{equation}
where \(\beta_s>0\) is a scale factor. 
This term explicitly regularizes the model toward source semantics and prevents the decision boundary from drifting away from the source class structure during test-time adaptation.

\emph{(ii) Mixup replay.}
MixUp~\cite{zhang2017mixup} and its variants are known to improve robustness by encouraging approximately linear behavior between samples, which effectively smooths decision boundaries.
We tailor this idea to our anchor-based setting by mixing each target sample with its \emph{matched} anchor, rather than mixing arbitrary pairs.
This yields a semantically meaningful interpolation path that connects a shifted target input to its corresponding source reference, providing targeted regularization where adaptation is most required.

Concretely, given a target--anchor pair \((x_b^t,\tilde{x}_b)\), we form the mixed input
\begin{equation}\label{eq:replay_mixup_x}
x^{m}_b = \lambda x^t_b + (1-\lambda)\tilde{x}_b, 
\qquad 
\lambda\sim\mathrm{Beta}(\alpha,\alpha),
\end{equation}
and the corresponding mixed soft target
\begin{equation}\label{eq:replay_mixup_p}
p^m_b = \lambda\,\mathrm{softmax}\!\big(f_{\theta}(x^t_b)\big)
+(1-\lambda)\,\mathrm{softmax}\!\big(\tilde{s}_b\big).
\end{equation}
We then enforce prediction consistency on the mixed sample:
\begin{equation}\label{eq:replay_mixup}
\mathcal{L}_{\text{mix}} = \mathrm{KL}\!\left(
\mathrm{softmax}\!\big(f_{\theta}(x^m_b)\big)\,
\middle\|\,
p^m_b
\right).
\end{equation}
This manifold replay encourages the decision boundary to vary smoothly along the target--anchor direction, mitigating local overfitting and improving robustness under continuous shift.
Finally, we combine the two components:
\begin{equation}\label{eq:replay_total}
\mathcal{L}_{\text{replay}}=\mathcal{L}_{\text{dir}}+\mathcal{L}_{\text{mix}}.
\end{equation}

\begin{table}[t]
\begin{center}
\setlength{\tabcolsep}{3pt}
\caption{
Experiments on ROID \cite{marsden2024universal} evaluate MixUp-style variants (\textbf{T-T}: Target-Target; \textbf{A-A}: Anchor-Anchor; \textbf{T-A}: Target-Anchor), reporting classification error (\%).
}
\vspace{-9pt}
\resizebox{0.60\linewidth}{!}{
\begin{tabular}{c|c|ccc}
\toprule
Baseline & DO-ALL without MixUp & T-T & A-A & T-A  \\
\midrule
54.40 & 53.77 & 53.68 & 53.69 & \textbf{53.55} \\
\bottomrule
\end{tabular}
}
\vspace{-25pt}
\label{tab:abl_mixup}
\end{center}
\end{table}

To validate the importance of mixing each target sample with its matched anchor, Table~\ref{tab:abl_mixup} compares three mixing variants.
While Target-Target and Anchor-Anchor mixing provide modest gains over the baseline, our Target-Anchor mixing yields the largest improvement, supporting the intuition that replay and smoothing are most effective when applied along the semantically meaningful path connecting each target to its distilled source anchor.

\noindent\textbf{(2) Feature-level alignment:}
While anchor soft labels preserve decision-level semantics, they do not explicitly constrain the feature-space geometry during adaptation.
As the target stream evolves, the feature extractor $g_\theta$ may drift even if predictions remain consistent, leading to representation collapse or overfitting to target domains.
To maintain the source manifold structure, we align anchor and target features in the latent space using a layer-wise MMD objective:
\begin{equation}\label{eq:mmd_loss}
\mathcal{L}_{\text{mmd}}
=\sum_{\ell\in{L}}
\big\|
\mathbb{E}_{x^t}[\,\phi(g_{\theta}^{\ell}(x^t))\,]
-\mathbb{E}_{\tilde{x}}[\,\phi(g_{\theta}^{\ell}(\tilde{x}))\,]
\big\|_{\mathcal{H}}^{2},
\end{equation}
where $\phi(\cdot)$ denotes the feature mapping to the reproducing kernel Hilbert space.
This encourages the evolving representation to remain aligned with the anchor space and reduces representation drift across layers.

\noindent\textbf{Total loss:}
To sum up, the total adaptation loss is:
\begin{equation}\label{eq:total_loss}
\mathcal{L}_{\text{anchor}}
= \mathcal{L}_{\text{replay}}
+ \lambda_{\text{mmd}}\,\mathcal{L}_{\text{mmd}},
\end{equation}
where $\lambda_{\text{mmd}}$ balances alignment and smoothness.
In short, $\mathcal{L}_{\text{replay}}$ preserves source competence, $\mathcal{L}_{\text{mmd}}$ maintains representational structure, and $\mathcal{L}_{\text{mix}}$ stabilizes local decision geometry—together enabling stable and robust CTTA.

\subsection{Harm-Adaptive Blending with Anchors}
\label{sec:HAB}
While the associated anchors guide the adaptation direction, the model parameters can still accumulate harmful drift over time, especially under long, noisy test streams.
To preserve the source-aligned representation encoded by the distilled anchors, we introduce a harm-adaptive blending mechanism that selectively pulls unstable parameter groups back toward the source model, while keeping beneficial updates intact.

After each anchor-based update, we compute the parameter change \(\Delta\theta=\theta_{\text{new}}-\theta_{\text{prev}}\) and the anchor-driven gradient \(g_S=\nabla_{\theta}\mathcal{L}_{\text{anchor}}\). 
Parameters are divided into groups \(\{G\}\) (\textit{e.g.}, by layer or parameter), and each group’s harm score is measured as:
\begin{equation}
    \mathrm{score}(G)
=\sum_{p\in G}\!\left(\langle g_S^{(p)},\Delta\theta^{(p)}\rangle
+\tfrac{1}{2}\hat{h}^{(p)}(\Delta\theta^{(p)})^{2}\right),
\end{equation}
where \(\hat{h}\) is the running average of squared gradients. 

Groups whose scores exceed a threshold $\rho$ are partially reverted to the initial source model $f_{\theta_0}$:
\begin{equation}
    \theta^{(p)}\!\leftarrow\!(1-\beta_G)\theta^{(p)}+\beta_G\theta_{0}^{(p)},
\end{equation}
where \(\beta_G=\beta_{\max}\sigma(\beta_{s}(s_G-0.5))\) scaled with the normalized harm score \(s_G\!\in[0,1]\). 
This blending mitigates harmful updates while retaining benefits from the distilled anchors, resulting in a stable yet flexible CTTA.

\section{Experiments}\label{sec:exp}

\input{tab/imagnet_comparison}

\subsection{Settings}
\noindent\textbf{Datasets:}
We conduct extensive experiments to evaluate the robustness of our method under CTTA scenarios. 
The primary evaluation uses CIFAR100-C and ImageNet-C~\cite{hendrycks2019benchmarking}, which extend CIFAR-100~\cite{krizhevsky2009learning} and ImageNet~\cite{deng2009imagenet} with distribution shifts. 
Each benchmark comprises fifteen corruption types and five severity levels. 
Results are reported at severity level five. 
On these benchmarks, we perform CTTA by adapting online to a sequence of fifteen corruptions, each presented as a stream of 5{,}000 unlabeled target images, following common practice~\cite{dobler2023robust, ni2025maintaining, marsden2024universal}.
We further assess scalability on the CCC benchmark~\cite{press2023rdumb}. 
CCC is designed to probe extreme-length adaptation on a continuously evolving stream of 7.5 million images with three severity levels. 

\noindent\textbf{Implementation:}
Following prior works~\cite{dobler2023robust,ni2025maintaining,marsden2024universal}, we employ ResNeXt-29~\cite{xie2017aggregated} for CIFAR100-C and ResNet-50~\cite{he2016deep} for ImageNet-C and CCC benchmark.
Hyperparameters are set to $\lambda_{\mathrm{mmd}}=10$, $\beta_{\max}=0.05$, and $\beta_{s}=5$ across all experiments.
Unless otherwise noted, anchors are distilled with WMDD~\cite{liu2023dataset} using the same backbone as the source model to preserve feature compatibility.
We also evaluate anchors distilled with DELT~\cite{shen2025delt} and SRe2L~\cite{yin2023squeeze}, observing comparable effectiveness.
Interestingly, we find that anchors distilled with a backbone different from the one used for CTTA remain equally effective, implying that DO-ALL can be generalized across architectures.

\subsection{Performance Evaluation}
\label{sec:main}
\subsubsection{Imagenet-C}
We first evaluate DO-ALL with three representative CTTA baselines, including EATA~\cite{niu2022efficient}, RMT~\cite{dobler2023robust}\footnote{In fact, RMT itself stores source prototypes and leverages them for CTTA. Our experimental results show that incorporating information from DD through the proposed method can still yield meaningful gains even on top of this approach.}, and ROID~\cite{marsden2024universal}.
As DO-ALL is designed in a plug-and-play manner, it can be seamlessly integrated into the conventional algorithms, without modifying the adaptation objective or the architecture of these methods.
The distilled anchor set is carried with the model and referenced during adaptation.
Across all evaluations, we observe consistent improvements regardless of the baseline, indicating that DO-ALL provides complementary advantages for the existing CTTA algorithms.

Table~\ref{tab:quan_in} reports results on ImageNet-to-ImageNet-C under the highest corruption severity (level 5) in an online adaptation setting.
DO-ALL consistently improves each base method across nearly all corruption types.
For example, EATA+DO-ALL reduces the average error from 58.0\% to 56.6\% and RMT+DO-ALL from 59.8\% to 57.4\%.
Given that severity level 5 induces highly unstable adaptation dynamics, these improvements highlight DO-ALL’s ability to anchor representation updates, preventing drift and ensuring more stable long-horizon adaptation.
The Results on CIFAR100-C are in Supplementary Material.

\vspace{-10pt}

\begin{figure}[t]
  \centering
  \begin{minipage}[t]{0.60\linewidth}
    \centering
    \captionof{table}{Classification accuracy (\%) on CCC benchmark. All results are averaged over 9 runs (3 seeds per each transition speed). $\dagger$ denotes reproduced results. \textbf{Bold} indicates the best results.}
    \vspace{3pt}
    \resizebox{\linewidth}{!}{
    \input{tab/CCC_comparison}
    }
  \end{minipage}\hfill
  \begin{minipage}[t]{0.38\linewidth}
    \centering
    \captionof{table}{Ablation study for DO-ALL. The results report the classification error rate (\%) on ImageNet-to-ImageNet-C. The baseline method is ROID~\cite{marsden2024universal}. \textbf{Bold} indicates the best results.}
    \vspace{3pt}
    \resizebox{\linewidth}{!}{

\input{tab/module_ablation}
    }
  \end{minipage}
  \vspace{-3pt}
\end{figure}

\subsubsection{CCC Benchmark}
\vspace{-5pt}
We further evaluate DO-ALL under the CCC benchmark, which is designed to measure stability under long-horizon, continuously accumulating distribution shifts. 
Unlike corruption benchmarks such as ImageNet-C and CIFAR100-C, where shifts occur independently per sample, CCC gradually transitions the test stream across domains, making models highly vulnerable to representation drift. 
As shown in Table~\ref{tab:quan_ccc}, ROID benefits noticeably from the introduction of DO-ALL. 
While ROID achieves an average accuracy of 33.6\%, DO-ALL improves this to 34.7\%, with the significant gain in the most important Hard  scenario (13.2\% to 15.5\%). 
This setting is particularly challenging due to the prolonged exposure to unseen distributions, and the improvement confirms that the distilled anchors serve as a persistent reference that prevents collapse over long periods. 

DO-ALL not only improves short-term adaptation quality but also delivers robust long-term stability in continuously evolving environments, confirming the claim of our \textbf{DO-ALL} framework: \textbf{Distill Once, Adapt Life-Long}.

\subsection{Ablation Study}

\subsubsection{Components Analysis}
\vspace{-5pt}
Table~\ref{tab:abl_component} presents the ablation results of the four components in DO-ALL.
Introducing each component individually (Rows A$\sim$C) yields only small but consistent reductions in error, indicating that Replay, MMD, and MixUp each contribute non-overlapping stabilization effects.
Specifically, Replay (A) slightly reduces forgetting, MMD alignment (B) mitigates representation drift, and MixUp smoothing (C) regularizes local adaptation behavior.
Because these components operate on different aspects of the adaptation process, their contributions do not conflict.

When combined (Row D), the improvement becomes clearer, reducing the average error from 54.5\% to 54.0\%.
Finally, harm-adaptive Blending (E, DO-ALL) yields the best performance of 53.6\%, confirming that preventing harmful parameter drift accumulated over long adaptation trajectories is essential to stable CTTA.
The results show that the components are complementary, and their combination is necessary for the full benefit of DO-ALL.

\subsubsection{Anchor Association Strategy}
\vspace{-5pt}
While DD provides high-quality anchors, their effectiveness depends critically on how they are used during adaptation.
Table~\ref{tab:abl_selection} compares different strategies for assigning anchors to target samples.
Assigning the nearest anchor in feature space consistently yields the best accuracy across all baselines.
In contrast, selecting anchors at random provides negligible benefit, and selecting the farthest anchor even degrades performance (\textit{e.g.}, RMT: 59.8\% → 61.9\%).
This confirms that DO-ALL’s stabilization does not arise merely from replaying DD-based anchors, but specifically from preserving semantic correspondence between shifted target and their counterparts.

\begin{figure}[t]
  \centering
  \begin{minipage}[t]{0.44\linewidth}
    \centering
    \captionof{table}{Ablation study of the anchor association method. The results report the classification error rate (\%) on ImageNet-to-ImageNet-C, averaged over 5 runs. All experiments are conducted with IPC=10 setting. \textbf{Bold} indicates the best results.}
    \vspace{5pt}
    \resizebox{\linewidth}{!}{
    \input{tab/selection_ablation}
    }
  \end{minipage}\hfill
  \begin{minipage}[t]{0.54\linewidth}
    \centering
    \captionof{table}{Accuracy--efficiency trade-off on ImageNet-to-ImageNet-C.
We report classification error rate (\%) with FPS, peak GPU memory,
and anchor storage on CPU. Experiments are conducted using anchors with IPC $=10$.}
\vspace{3pt}
    \resizebox{\linewidth}{!}{
    \begin{tabular}{l|c|c|c|c}
\toprule
\multirow{2}{*}{Method}  & \multirow{2}{*}{FPS} & GPU Mem & Anchor Size & Err. \\
& & (GB) & (GB) & (\%) \\
\midrule
Baseline (ROID~\cite{marsden2024universal})              & 284.9 & 9.32  & N/A    & 54.5 \\
+DO-ALL (stride-1)           & 106.3 & 15.94 & 1.55   & 53.6 \\
+DO-ALL (stride-3)           & 187.5 & 15.94 & 1.55   & 53.7 \\
+DO-ALL (stride-5)           & 220.4 & 15.94 & 1.55   & 53.8 \\
+DO-ALL (stride-7)           & 244.0 & 15.94 & 1.55   & 53.9 \\
\bottomrule
\end{tabular}
\label{tab:efficiency}
    }
  \end{minipage}
  \vspace{-10pt}
\end{figure}

\subsection{Efficiency}
DO-ALL augments a base CTTA method with anchor-guided objectives and harm-adaptive blending, while leaving the original CTTA pipeline unchanged.
Importantly, the distilled anchor set is generated \emph{once} before deployment (offline distillation) and carried alongside the model at test time.
As a result, DO-ALL introduces no additional distillation cost during deployment and can be plugged into existing CTTA algorithms as a lightweight stabilization component.

To further reduce the overhead during CTTA, we test to compute the anchor branch only every $k$ test-time updates (\emph{stride-$k$}).
At step $t$, if $t \bmod k \neq 0$, we optimize only the base CTTA objective; if $t \bmod k = 0$, we additionally apply $L_{\text{anchor}}$ and perform harm-adaptive blending.
This amortizes the anchor computation across updates without altering the core CTTA optimization.
As shown in Table~\ref{tab:efficiency}, DO-ALL consistently improves robustness over the ROID baseline even with large strides (e.g., stride-7), demonstrating a favorable accuracy--efficiency trade-off.
We additionally report frames per second (FPS) and peak GPU memory (GB) on a single RTX 3090 using identical batch size and data-loading settings across methods.
We also include the host-side (CPU) storage required for the anchor bank.
Table~\ref{tab:efficiency} summarizes the resulting trade-off in terms of error, FPS, GPU memory, and CPU anchor storage.

Overall, DO-ALL incurs only modest test-time overhead and requires a small host-side anchor storage, while delivering consistent robustness gains. 
This suggests that DO-ALL can be an attractive direction for CTTA.


\subsection{In-Depth Analysis on Source-Distilled Anchors}





\begin{figure}[t]
  \centering
  \begin{minipage}[t]{0.48\linewidth}
    \centering
    \captionof{table}{Classification error rate (\%) on ImageNet-to-ImageNet-C using various DD methods. All results are evaluated with the largest corruption severity level 5 in an online manner.}
    \resizebox{\linewidth}{!}{

\input{tab/coreset_and_DD}
    }
  \end{minipage}\hfill
  \begin{minipage}[t]{0.48\linewidth}
    \centering
    \captionof{table}{Impact of images per class~(IPC) on DO-ALL. The
results report the classification error rate (\%) on ImageNet-to-
ImageNet-C, over 5 runs.}
    \resizebox{\linewidth}{!}{
    \input{tab/ipc_ablation}
    }
  \end{minipage}
\end{figure}

\subsubsection{Experiments on Diverse DD Methods}
To examine whether DO-ALL relies on a specific dataset distillation algorithm, we also evaluate its performance using several recent DD methods: SRe2L~\cite{yin2023squeeze}, DELT~\cite{shen2025delt}, and WMDD~\cite{liu2023dataset}.
Results are shown in Table~\ref{tab:abl_coresetDD}.
Across all three CTTA baselines, every DD variant consistently improves performance over the baseline.
For example, ROID+DO-ALL improves from 54.5\% to 53.7\%, 53.6\%, and 53.6\% when using SRe2L, DELT, and WMDD, respectively.
These results show that DO-ALL does not depend on the choice of DD algorithm.
While different DD algorithms vary in how they construct synthetic samples, DO-ALL is able to effectively utilize the distilled set regardless of the specific DD objective or formulation.
\vspace{-5pt}

\subsubsection{Analysis on IPC}
\vspace{-5pt}
We further examine how the number of distilled samples per class (IPC) influences performance (Table~\ref{tab:abl_ipc}).
Across all CTTA baselines, introducing even a very small anchor set (IPC = 1) already yields noticeable improvements over the baseline, indicating that DO-ALL does not require a large number of anchors to be effective—a lightweight distilled set is sufficient to stabilize adaptation.
Additionally, increasing IPC provides generally consistent gains, reflecting that richer distilled information can further enhance stability.
DO-ALL allows users to flexibly choose the IPC that best fits their application, balancing performance improvements against the size of source-distilled anchors.
\vspace{-10pt}

\subsubsection{Impact of DD Performance on CTTA Gain}
\begin{figure}[t]
    \centering    \includegraphics[width=0.99\linewidth]{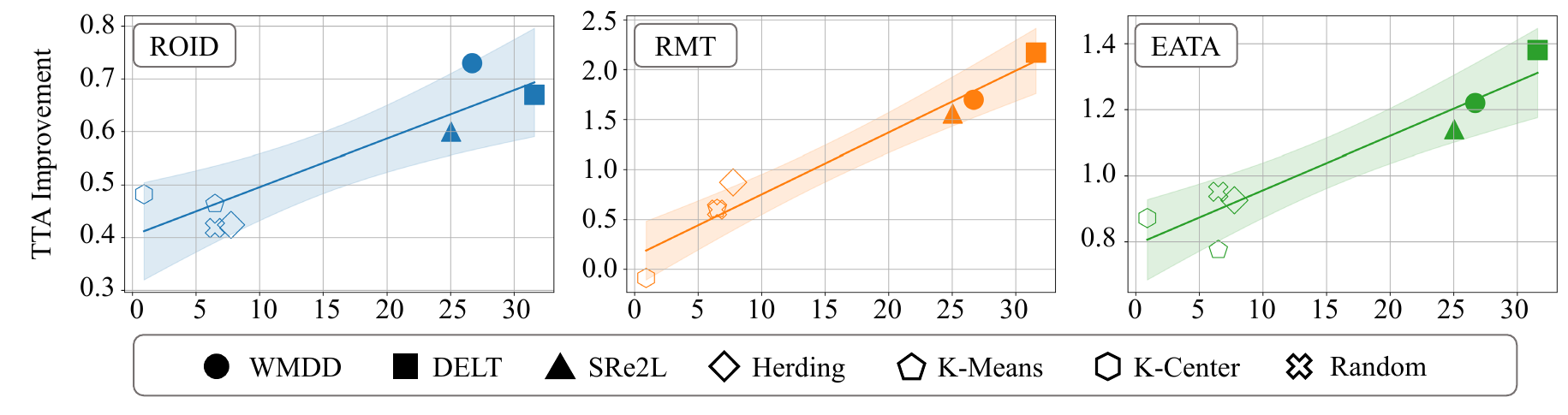}
    \vspace{-10pt}
    \caption{Correlation between the quality of the anchor set and its effect on CTTA. The x-axis denotes the validation accuracy obtained by training a classifier solely on the small set, while the y-axis shows the corresponding improvement achieved by integrating the set into DO-ALL. A positive trend indicates that higher-quality anchors lead to stronger CTTA stabilization, highlighting DO-ALL’s ability to exploit semantic richness in distilled data.}
    \label{fig:correlation}
\end{figure}

\vspace{-5pt}
To further analyze the relationship between the quality of the distilled (or coreset) anchors and their impact within DO-ALL, we plot in Fig.~\ref{fig:correlation} the validation performance of each small set (x-axis) against the resulting TTA improvement (y-axis).
Here, the validation performance refers to the accuracy obtained when a classifier is trained only on the small distilled or coreset set and then evaluated on the original source validation data.
This metric effectively reflects how well the small set preserves the semantic structure of the source distribution and is widely regarded as the standard evaluation criterion in the DD literature.

Interestingly, we observe a positive correlation across all CTTA baselines: anchor sets that achieve higher standalone validation accuracy also yield larger improvements when used within DO-ALL.
Notably, using dataset-distilled anchors (filled markers) consistently delivers larger TTA gains than coreset-selected anchors (hollow markers) across all CTTA methods.
This indicates that DO-ALL’s gains are not merely due to retaining a subset of source information, but rather stem from the structured, optimization-driven compression enabled by dataset distillation.
Moreover, the result suggests that higher semantic fidelity of the distilled set translates into stronger stabilization during CTTA, leading to more robust adaptation.
In other words, as DD methods continue to advance in capturing richer and more compact source semantics, DO-ALL can readily transfer these gains into more stable and robust CTTA performance by leveraging distilled anchors as high-density semantic cues.

Together with the earlier results demonstrating compatibility across diverse CTTA methods (Sec.~\ref{sec:main}), this correlation strongly suggests DO-ALL’s role as a plug-and-play bridge that seamlessly connects and benefits from progress in both fields of DD and CTTA.
\vspace{-5pt}

\subsubsection{Cross-Architecture Experiments}
\vspace{-5pt}
To further assess the flexibility of DO-ALL, we study whether distilled anchors remain effective when the architecture used for distillation differs from that of the CTTA model.
As shown in Table~\ref{tab:cross-architecture}, we distill anchors with ResNet-18 (RN18) and apply them to CTTA with ResNet-50 (RN50).
Across all baselines (ROID, RMT, and EATA), performance is virtually unchanged compared to using anchors distilled from the same backbone, indicating strong cross-backbone transfer.
We also evaluate ViT-based anchors and cross-family transfer (Table~\ref{tab:vit}).
Specifically, we distill anchors with ViT-B and test them on ResNet-50 CTTA (ViT$\rightarrow$CNN), and we further consider both matched (ViT$\rightarrow$ViT) and reverse transfer (CNN$\rightarrow$ViT) settings.
In all cases, DO-ALL consistently improves over the corresponding baseline.
Overall, these results suggest that DO-ALL does not require architectural alignment between the distillation and adaptation stages.

\begin{figure}[t]
  \centering
  \begin{minipage}[t]{0.49\linewidth}
    \centering
    \captionof{table}{Experiments with data distilled using different architectures (ResNet-18) on ImageNet-to-ImageNet-C over 5 runs.}
    \resizebox{\linewidth}{!}{
    \input{tab/diverseDD_arch}
    }
  \end{minipage}\hfill
  \begin{minipage}[t]{0.49\linewidth}
    \centering
    \captionof{table}{Experiments on ImageNet-C with a ViT architecture. ROID is used as the baseline CTTA method. For experiments with LGM, we set IPC to 1.}
    \resizebox{\linewidth}{!}{
    \begin{tabular}{c|c|cc}
\toprule
   DD Method & TTA Arch.  & Baseline&  DO-ALL \\
 \midrule
 LGM~\cite{cazenavette2025dataset} with ViT-B & RN50 & 54.5 & 53.8 \\
 LGM~\cite{cazenavette2025dataset} with ViT-B & ViT-B & 45.0 & 44.5 \\
 WMDD~\cite{liu2023dataset} with RN50 & ViT-B  & 45.0 & 44.4 \\
\bottomrule
\end{tabular}
\label{tab:vit}
    }
  \end{minipage}
\end{figure}



\section{Conclusion}
\label{sec:conclusion}
\vspace{-5pt}
In this work, we explore the use of dataset distillation as a compact mechanism for retaining source-domain knowledge in continual test-time adaptation.
By leveraging a compact set of source-distilled anchors and a lightweight integration strategy, DO-ALL offers a plug-and-play stabilization layer for CTTA that works across different adaptation algorithms and different distillation methods, without modifying the underlying model.
Our experiments demonstrate that DO-ALL meaningfully improves long-term adaptation stability across multiple datasets and settings, while offering a flexible performance–memory trade-off.
In addition, DO-ALL is orthogonal to most test-time objectives and can be combined with future advances in both CTTA and dataset distillation, making it a practical pathway for continual deployment under strict privacy and storage constraints.
We believe a promising direction is to further improve anchor selection and update schedules under non-stationary streams, and to study robustness when the target shift is abrupt or adversarially corrupted.
More broadly, our findings suggest that distilled data can function as a general-purpose, compact memory to stabilize online learning beyond CTTA.

\section{Acknowledgement}

This work was supported by the National Research Foundation of Korea(NRF) grant funded by the Korea government(MSIT) (RS-2026-25473963, RS-2026-25478915).



%
%
\bibliographystyle{splncs04}
\bibliography{main}
\end{document}

%% file: tab/imagnet_comparison.tex
\definecolor{rgrey}{RGB}{250,244,244}  
\definecolor{ggrey}{RGB}{243,250,246}  
\definecolor{bgrey}{RGB}{242,246,250}  
\definecolor{ygrey}{RGB}{250,250,243}  

\begin{table*}[t]
\begin{center}
\caption{Classification error rate (\%) on ImageNet-to-ImageNet-C. All results
are evaluated with the largest corruption severity level 5 in an online manner. 
We report the performance of our method averaged over 5
runs.}
\vspace{-12pt}
\resizebox{\textwidth}{!}{
\begin{tabular}{l|ccccc ccccc ccccc|c}
\toprule
Method & Gau. & shot & imp. & def. & glass & mot. & zoom & snow & fro. & fog & bri. & con. & ela. & pix. & jpeg & Avg.\\ \midrule
Source & 97.8 & 97.1 & 98.2&  81.7 & 89.8 & 85.2 & 78.0&  83.5& 77.1& 75.9&  41.3 & 94.5&  82.5& 79.3& 68.6 & 82.0 \\
TENT$_{\textit{ICLR21}}$\cite{wang2020tent} &81.6& 74.6& 72.7& 77.6& 73.8&  65.5&  55.3&  61.6&  63.0&  51.7&  38.2&  72.1&  50.8&  47.4&  53.3&  62.6 \\
CoTTA$_{\textit{CVPR22}}$\cite{wang2022continual}  & 84.7& 82.1& 80.6& 81.3& 79.0& 68.6& 57.5& 60.3& 60.5& 48.3 &36.6& 66.1 &47.3 &41.2& 46.0& 62.7 \\
DSS$_{\textit{WACV24}}$\cite{wang2024continual} & 82.3 &78.4 &76.7& 81.9& 77.8 &66.9& 60.9& 50.8& 60.9& 47.7& 35.4 &69.0& 47.5& 40.9 &46.2 &62.2 \\
BeCoTTA$_{\textit{ICML24}}$\cite{lee2024becotta} &  84.1& 74.3& 72.2& 77.4& 71.9 &63.4& 55.1 &57.2 &61.2& 50.7& 36.4& 66.1& 49.2 &45.6 &48.4& 60.9 \\
OBAO$_{\textit{ECCV24}}$\cite{zhu2024reshaping}  & 78.5 & 75.3& 73.0& 75.7& 73.1& 64.5& 56.0& 55.8& 58.1& 47.6 &38.5& 58.5 &46.1 &42.0& 43.4& 59.0 \\
TCA$_{\textit{CVPR25}}$\cite{ni2025maintaining} & 78.3  &71.8 & 73.5  &74.4 & 73.5 & 63.3 & 56.5 & 56.9  &59.4  &48.1  &39.6 & 59.6 & 47.2 & 42.9 & 44.7 & 59.3 \\
\rowcolor{rgrey}EATA$_{\textit{ICML22}}$\cite{niu2022efficient} & 76.3 & 66.5 & 65.0 & 73.1 & 69.1 & 62.1 & 53.5 & 58.9 & 59.3 & 48.1 & 35.9 & 62.8 & 47.5 &43.9 & 47.5 & 58.0$\pm$0.18 \\
\rowcolor{ggrey}RMT$_{\textit{CVPR23}}$\cite{dobler2023robust}  &77.9 & 73.1& 69.9& 73.5&71.1 &63.1 & 57.1& 57.1& 59.2& 50.4 & 42.9& 60.1&49.0 &45.7 &46.9 &59.8$\pm$0.18\\
\rowcolor{bgrey}ROID$_{\textit{WACV24}}$\cite{marsden2024universal} & 71.7& 62.2 &62.2& 69.6& 66.5 &57.1 &49.3 &52.3& 57.4 &43.5& 33.4 &59.1 &45.4 &41.8 &46.2 &54.5$\pm$0.10 \\
\rowcolor{ygrey}ASR$_{\textit{ICLR26}}$\cite{lim2026and} & 70.5 & 61.3 & 61.0 & 69.2 & 66.5 & 57.3 & 49.9 & 52.5 & 57.3 & 44.5 & 34.3 & 57.4 & 45.2 & 42.1 & 45.2 & 54.3$\pm$0.14 \\
\midrule
\rowcolor{rgrey}EATA + DO-ALL & 73.2 &	63.3&	62.7&	73.0 &	67.8 &	61.3 &	53.0 &	56.4 &	58.1 &	46.6 &	34.9 &	61.8 &	47.4 &	42.6 &	46.3 &	56.6$\pm$0.22 \\ 
\rowcolor{ggrey}RMT + DO-ALL & 75.6 &	70.6& 69.0 &	72.7&	70.4&	61.8&	54.4&	54.0&	57.2&	46.0&	39.0&	56.9&	45.9&	42.7&	44.4&	57.4$\pm$0.23 \\ 
\rowcolor{bgrey}ROID + DO-ALL & 68.9 &60.5&	60.5 &	69.2 &	65.8&	56.6 &	48.4&	51.5&	56.9 &	42.3&	32.8&	58.9&	44.6&	41.0&	45.4 & 53.6$\pm$0.09 \\ 
\rowcolor{ygrey}ASR + DO-ALL  & 69.3 & 59.9 & 60.6 & 69.1 & 65.6 & 57.0 & 49.4 & 52.1 & 56.8 & 43.7 & 33.7 & 57.3 & 44.7 & 41.2 & 44.8 & 53.7$\pm$0.13\\ \bottomrule
\end{tabular}
}
\vspace{-15pt}
\label{tab:quan_in}
\end{center}
\end{table*}

%% file: tab/CCC_comparison.tex

\begin{tabular}{l|ccc|c}\toprule
Method & CCC-Easy & CCC-Medium &CCC-Hard & Avg. \\
 \midrule
CoTTA~\cite{wang2022continual}& 14.9$\pm$0.88 & 7.7$\pm$0.43 & 1.1$\pm$0.16 & 7.9 \\
ETA~\cite{niu2022efficient} & 41.4$\pm$0.95 & 1.1$\pm$0.43 & 0.2$\pm$0.05 & 14.2\\
EATA~\cite{niu2022efficient}  & 48.2$\pm$0.60 & 35.4$\pm$1.02 & 8.7$\pm$0.80 & 30.8 \\
SANTA~\cite{chakrabarty2023santa} & 47.8$\pm$0.46 & 32.7$\pm$0.80 & 9.1$\pm$0.60 & 29.9\\
RDumb~\cite{press2023rdumb} & \textbf{49.3$\pm$0.88} & 38.9$\pm$1.40 & 9.6$\pm$1.60 & 32.6 \\
TCA~\cite{ni2025maintaining}   & 49.1$\pm$0.35 & 39.5$\pm$0.53 & 10.1$\pm$0.22 & 32.9 \\
\rowcolor{grey1} ROID$^\dagger$~\cite{marsden2024universal}   & 48.5$\pm$0.63 & 39.2$\pm$1.98 & 13.2$\pm$2.47 & 33.6  \\ \midrule
\rowcolor{grey1} ROID+DO-ALL  &48.9$\pm$0.64 &\textbf{39.6$\pm$1.97} &\textbf{15.5$\pm$1.49}  &\textbf{34.7}\\ 
\bottomrule
\end{tabular}
\label{tab:quan_ccc}

%% file: tab/module_ablation.tex
\begin{tabular}{c|cccc|c}
\toprule
Exp. & $\mathcal{L}_{\text{dir}}$ & $\mathcal{L}_{\text{mix}}$ & $\mathcal{L}_{\text{mmd}}$  & Blend & Avg. \\ 
\midrule
Baseline & & & & & 54.5 \\
A & \checkmark & & & &54.1  \\
B &  & \checkmark & &  &54.1   \\
C &  & &\checkmark & &  54.2 \\ 
D & \checkmark& \checkmark& \checkmark& &  54.0 \\ 
E & \checkmark& \checkmark& \checkmark& \checkmark& \textbf{53.6}  \\ 
\bottomrule
\end{tabular}
\label{tab:abl_component}

%% file: tab/selection_ablation.tex


\begin{tabular}{l|c|ccc}
\toprule
 CTTA & Baseline & Random & Farthest & Nearest  \\
\midrule
EATA~\cite{niu2022efficient} & 58.0 & 57.3 & 58.7  & \textbf{56.6}    \\
RMT~\cite{dobler2023robust} & 59.8 & 60.2 & 61.9  & \textbf{57.4}    \\
ROID~\cite{marsden2024universal} & 54.5 & 54.0 & 54.0 & \textbf{53.6}     \\

\bottomrule
\end{tabular}
\label{tab:abl_selection}

%% file: tab/coreset_and_DD.tex
\begin{tabular}{l|c|ccc}
\toprule
CTTA & Baseline & SRe2L & DELT & WMDD \\
\midrule
 EATA~\cite{niu2022efficient}  & 58.0 & 56.6 & 56.4 & 56.6  \\
 RMT~\cite{dobler2023robust}  & 59.8 & 57.5 & 56.9 & 57.4 \\
 ROID~\cite{marsden2024universal}  & 54.5 & 53.7 & 53.6 & 53.6  \\
\bottomrule
\end{tabular}
\label{tab:abl_coresetDD}

%% file: tab/ipc_ablation.tex

\begin{tabular}{l|c|ccc}
\toprule
 CTTA & Baseline & IPC=1 & IPC=5 & IPC=10  \\
\midrule
EATA~\cite{niu2022efficient} & 58.0 & 57.8 & 56.7 & 56.6  \\
RMT~\cite{dobler2023robust}  & 59.8 & 57.9 & 57.6 & 57.4  \\
ROID~\cite{marsden2024universal} & 54.5 & 53.7 & 53.6 & 53.6   \\
\bottomrule
\end{tabular}
\label{tab:abl_ipc}


%% file: tab/diverseDD_arch.tex


\begin{tabular}{l|c|ccc}
\toprule
 CTTA & Baseline & SRe2L & DELT & WMDD \\
\midrule
ROID~\cite{marsden2024universal} &54.5  & 53.7 & 53.7 & 53.7  \\
RMT~\cite{dobler2023robust} & 59.1 & 57.8  & 58.1 & 57.8  \\
EATA~\cite{niu2022efficient} &57.8& 56.8 & 56.6  & 56.7  \\

\bottomrule
\end{tabular}
\label{tab:cross-architecture}

%% file: main.bib
@String(ICLR  = {Int. Conf. Learn. Represent.})

@String(AAAI  = {AAAI})

@String(ICIP  = {IEEE Int. Conf. Image Process.})

@String(ICLR  = {ICLR})

@String(ICIP  = {ICIP})

@article{wang2018dataset,
  title={Dataset distillation},
  author={Wang, Tongzhou and Zhu, Jun-Yan and Torralba, Antonio and Efros, Alexei A},
  journal={arXiv preprint arXiv:1811.10959},
  year={2018}
}

@inproceedings{cazenavette2022dataset,
  title={Dataset distillation by matching training trajectories},
  author={Cazenavette, George and Wang, Tongzhou and Torralba, Antonio and Efros, Alexei A and Zhu, Jun-Yan},
  booktitle={Proceedings of the IEEE/CVF Conference on Computer Vision and Pattern Recognition},
  pages={4750--4759},
  year={2022}
}

@inproceedings{zhao2023dataset,
  title={Dataset condensation with distribution matching},
  author={Zhao, Bo and Bilen, Hakan},
  booktitle={Proceedings of the IEEE/CVF Winter Conference on Applications of Computer Vision},
  pages={6514--6523},
  year={2023}
}

@inproceedings{wang2022cafe,
  title={Cafe: Learning to condense dataset by aligning features},
  author={Wang, Kai and Zhao, Bo and Peng, Xiangyu and Zhu, Zheng and Yang, Shuo and Wang, Shuo and Huang, Guan and Bilen, Hakan and Wang, Xinchao and You, Yang},
  booktitle={Proceedings of the IEEE/CVF Conference on Computer Vision and Pattern Recognition},
  pages={12196--12205},
  year={2022}
}

@inproceedings{cui2023scaling,
  title={Scaling up dataset distillation to imagenet-1k with constant memory},
  author={Cui, Justin and Wang, Ruochen and Si, Si and Hsieh, Cho-Jui},
  booktitle={International Conference on Machine Learning},
  pages={6565--6590},
  year={2023},
  organization={PMLR}
}

@inproceedings{zhao2023improved,
  title={Improved distribution matching for dataset condensation},
  author={Zhao, Ganlong and Li, Guanbin and Qin, Yipeng and Yu, Yizhou},
  booktitle={Proceedings of the IEEE/CVF Conference on Computer Vision and Pattern Recognition},
  pages={7856--7865},
  year={2023}
}

@article{cui2022dc,
  title={Dc-bench: Dataset condensation benchmark},
  author={Cui, Justin and Wang, Ruochen and Si, Si and Hsieh, Cho-Jui},
  journal={Advances in Neural Information Processing Systems},
  volume={35},
  pages={810--822},
  year={2022}
}

@inproceedings{zhang2024m3d,
  title={M3d: Dataset condensation by minimizing maximum mean discrepancy},
  author={Zhang, Hansong and Li, Shikun and Wang, Pengju and Zeng, Dan and Ge, Shiming},
  booktitle={Proceedings of the AAAI Conference on Artificial Intelligence},
  volume={38},
  number={8},
  pages={9314--9322},
  year={2024}
}

@article{li2025dd,
  title={Dd-ranking: Rethinking the evaluation of dataset distillation},
  author={Li, Zekai and Zhong, Xinhao and Khaki, Samir and Liang, Zhiyuan and Zhou, Yuhao and Shi, Mingjia and Wang, Ziqiao and Zhao, Xuanlei and Zhao, Wangbo and Qin, Ziheng and others},
  journal={arXiv preprint arXiv:2505.13300},
  year={2025}
}

@article{zhao2020dataset,
  title={Dataset condensation with gradient matching},
  author={Zhao, Bo and Mopuri, Konda Reddy and Bilen, Hakan},
  journal={arXiv preprint arXiv:2006.05929},
  year={2020}
}

@inproceedings{zhao2021dataset,
  title={Dataset condensation with differentiable siamese augmentation},
  author={Zhao, Bo and Bilen, Hakan},
  booktitle={International Conference on Machine Learning},
  pages={12674--12685},
  year={2021},
  organization={PMLR}
}

@article{zhang2024dance,
  title={Dance: Dual-view distribution alignment for dataset condensation},
  author={Zhang, Hansong and Li, Shikun and Lin, Fanzhao and Wang, Weiping and Qian, Zhenxing and Ge, Shiming},
  journal={arXiv preprint arXiv:2406.01063},
  year={2024}
}

@inproceedings{wang2025dataset,
  title={Dataset distillation with neural characteristic function: A minmax perspective},
  author={Wang, Shaobo and Yang, Yicun and Liu, Zhiyuan and Sun, Chenghao and Hu, Xuming and He, Conghui and Zhang, Linfeng},
  booktitle={Proceedings of the Computer Vision and Pattern Recognition Conference},
  pages={25570--25580},
  year={2025}
}

@article{liu2023dataset,
  title={Dataset distillation via the wasserstein metric},
  author={Liu, Haoyang and Li, Yijiang and Xing, Tiancheng and Dalal, Vibhu and Li, Luwei and He, Jingrui and Wang, Haohan},
  journal={arXiv preprint arXiv:2311.18531},
  year={2023}
}

@article{shao2024elucidating,
  title={Elucidating the design space of dataset condensation},
  author={Shao, Shitong and Zhou, Zikai and Chen, Huanran and Shen, Zhiqiang},
  journal={Advances in neural information processing systems},
  volume={37},
  pages={99161--99201},
  year={2024}
}

@inproceedings{sun2024information,
  title={Information compensation: A fix for any-scale dataset distillation},
  author={Sun, Peng and Shi, Bei and Shang, Xinyi and Lin, Tao},
  booktitle={ICLR 2024 Workshop on Data-centric Machine Learning Research (DMLR)},
  year={2024}
}

@article{yin2023squeeze,
  title={Squeeze, recover and relabel: Dataset condensation at imagenet scale from a new perspective},
  author={Yin, Zeyuan and Xing, Eric and Shen, Zhiqiang},
  journal={Advances in Neural Information Processing Systems},
  volume={36},
  pages={73582--73603},
  year={2023}
}

@inproceedings{shen2025delt,
  title={Delt: A simple diversity-driven earlylate training for dataset distillation},
  author={Shen, Zhiqiang and Sherif, Ammar and Yin, Zeyuan and Shao, Shitong},
  booktitle={Proceedings of the Computer Vision and Pattern Recognition Conference},
  pages={4797--4806},
  year={2025}
}

@article{ma2025curriculum,
  title={Curriculum dataset distillation},
  author={Ma, Zhiheng and Cao, Anjia and Yang, Funing and Gong, Yihong and Wei, Xing},
  journal={IEEE Transactions on Image Processing},
  year={2025},
  publisher={IEEE}
}

@inproceedings{zhang2025infer,
  title={Breaking Class Barriers: Efficient Dataset Distillation via Inter-Class Feature Compensator},
  author={Zhang, Xin and Du, Jiawei and Liu, Ping and Zhou, Joey Tianyi},
  booktitle={Proceedings of the International Conference on Learning Representations (ICLR)},
  year={2025}
}

@article{yang2023efficient,
  title={An efficient dataset condensation plugin and its application to continual learning},
  author={Yang, Enneng and Shen, Li and Wang, Zhenyi and Liu, Tongliang and Guo, Guibing},
  journal={Advances in Neural Information Processing Systems},
  volume={36},
  pages={67625--67642},
  year={2023}
}

@inproceedings{dong2022privacy,
  title={Privacy for free: How does dataset condensation help privacy?},
  author={Dong, Tian and Zhao, Bo and Lyu, Lingjuan},
  booktitle={International Conference on Machine Learning},
  pages={5378--5396},
  year={2022},
  organization={PMLR}
}

@inproceedings{xiong2023feddm,
  title={Feddm: Iterative distribution matching for communication-efficient federated learning},
  author={Xiong, Yuanhao and Wang, Ruochen and Cheng, Minhao and Yu, Felix and Hsieh, Cho-Jui},
  booktitle={Proceedings of the IEEE/CVF Conference on Computer Vision and Pattern Recognition},
  pages={16323--16332},
  year={2023}
}

@inproceedings{song2023federated,
  title={Federated learning via decentralized dataset distillation in resource-constrained edge environments},
  author={Song, Rui and Liu, Dai and Chen, Dave Zhenyu and Festag, Andreas and Trinitis, Carsten and Schulz, Martin and Knoll, Alois},
  booktitle={2023 International Joint Conference on Neural Networks (IJCNN)},
  pages={1--10},
  year={2023},
  organization={IEEE}
}

@inproceedings{liang2020we,
  title={Do we really need to access the source data? source hypothesis transfer for unsupervised domain adaptation},
  author={Liang, Jian and Hu, Dapeng and Feng, Jiashi},
  booktitle={International conference on machine learning},
  pages={6028--6039},
  year={2020},
  organization={PMLR}
}

@article{li2016revisiting,
  title={Revisiting batch normalization for practical domain adaptation},
  author={Li, Yanghao and Wang, Naiyan and Shi, Jianping and Liu, Jiaying and Hou, Xiaodi},
  journal={arXiv preprint arXiv:1603.04779},
  year={2016}
}

@article{niu2023towards,
  title={Towards stable test-time adaptation in dynamic wild world},
  author={Niu, Shuaicheng and Wu, Jiaxiang and Zhang, Yifan and Wen, Zhiquan and Chen, Yaofo and Zhao, Peilin and Tan, Mingkui},
  journal={arXiv preprint arXiv:2302.12400},
  year={2023}
}

@article{zhang2022memo,
  title={Memo: Test time robustness via adaptation and augmentation},
  author={Zhang, Marvin and Levine, Sergey and Finn, Chelsea},
  journal={Advances in neural information processing systems},
  volume={35},
  pages={38629--38642},
  year={2022}
}

@article{wang2020tent,
  title={Tent: Fully test-time adaptation by entropy minimization},
  author={Wang, Dequan and Shelhamer, Evan and Liu, Shaoteng and Olshausen, Bruno and Darrell, Trevor},
  journal={arXiv preprint arXiv:2006.10726},
  year={2020}
}

@article{lim2023ttn,
  title={Ttn: A domain-shift aware batch normalization in test-time adaptation},
  author={Lim, Hyesu and Kim, Byeonggeun and Choo, Jaegul and Choi, Sungha},
  journal={arXiv preprint arXiv:2302.05155},
  year={2023}
}

@inproceedings{niu2022efficient,
  title={Efficient test-time model adaptation without forgetting},
  author={Niu, Shuaicheng and Wu, Jiaxiang and Zhang, Yifan and Chen, Yaofo and Zheng, Shijian and Zhao, Peilin and Tan, Mingkui},
  booktitle={International conference on machine learning},
  pages={16888--16905},
  year={2022},
  organization={PMLR}
}

@inproceedings{wang2022continual,
  title={Continual test-time domain adaptation},
  author={Wang, Qin and Fink, Olga and Van Gool, Luc and Dai, Dengxin},
  booktitle={Proceedings of the IEEE/CVF Conference on Computer Vision and Pattern Recognition},
  pages={7201--7211},
  year={2022}
}

@inproceedings{dobler2023robust,
  title={Robust mean teacher for continual and gradual test-time adaptation},
  author={D{\"o}bler, Mario and Marsden, Robert A and Yang, Bin},
  booktitle={Proceedings of the IEEE/CVF Conference on Computer Vision and Pattern Recognition},
  pages={7704--7714},
  year={2023}
}

@inproceedings{song2023ecotta,
  title={Ecotta: Memory-efficient continual test-time adaptation via self-distilled regularization},
  author={Song, Junha and Lee, Jungsoo and Kweon, In So and Choi, Sungha},
  booktitle={Proceedings of the IEEE/CVF Conference on Computer Vision and Pattern Recognition},
  pages={11920--11929},
  year={2023}
}

@inproceedings{yuan2023robust,
  title={Robust test-time adaptation in dynamic scenarios},
  author={Yuan, Longhui and Xie, Binhui and Li, Shuang},
  booktitle={Proceedings of the IEEE/CVF Conference on Computer Vision and Pattern Recognition},
  pages={15922--15932},
  year={2023}
}

@article{kang2023leveraging,
  title={Leveraging proxy of training data for test-time adaptation},
  author={Kang, Juwon and Kim, Nayeong and Kwon, Donghyeon and Ok, Jungseul and Kwak, Suha},
  year={2023}
}

@inproceedings{wang2024continual,
  title={Continual test-time domain adaptation via dynamic sample selection},
  author={Wang, Yanshuo and Hong, Jie and Cheraghian, Ali and Rahman, Shafin and Ahmedt-Aristizabal, David and Petersson, Lars and Harandi, Mehrtash},
  booktitle={Proceedings of the IEEE/CVF Winter Conference on Applications of Computer Vision},
  pages={1701--1710},
  year={2024}
}

@article{lee2024becotta,
  title={Becotta: Input-dependent online blending of experts for continual test-time adaptation},
  author={Lee, Daeun and Yoon, Jaehong and Hwang, Sung Ju},
  journal={arXiv preprint arXiv:2402.08712},
  year={2024}
}

@article{press2023rdumb,
  title={Rdumb: A simple approach that questions our progress in continual test-time adaptation},
  author={Press, Ori and Schneider, Steffen and K{\"u}mmerer, Matthias and Bethge, Matthias},
  journal={Advances in Neural Information Processing Systems},
  volume={36},
  pages={39915--39935},
  year={2023}
}

@inproceedings{adachi2023covariance,
  title={Covariance-aware feature alignment with pre-computed source statistics for test-time adaptation to multiple image corruptions},
  author={Adachi, Kazuki and Yamaguchi, Shin’Ya and Kumagai, Atsutoshi},
  booktitle={2023 IEEE International Conference on Image Processing (ICIP)},
  pages={800--804},
  year={2023},
  organization={IEEE}
}

@inproceedings{jung2023cafa,
  title={Cafa: Class-aware feature alignment for test-time adaptation},
  author={Jung, Sanghun and Lee, Jungsoo and Kim, Nanhee and Shaban, Amirreza and Boots, Byron and Choo, Jaegul},
  booktitle={Proceedings of the IEEE/CVF International Conference on Computer Vision},
  pages={19060--19071},
  year={2023}
}

@inproceedings{marsden2024universal,
  title={Universal test-time adaptation through weight ensembling, diversity weighting, and prior correction},
  author={Marsden, Robert A and D{\"o}bler, Mario and Yang, Bin},
  booktitle={Proceedings of the IEEE/CVF Winter Conference on Applications of Computer Vision},
  pages={2555--2565},
  year={2024}
}

@inproceedings{ni2025maintaining,
  title={Maintaining consistent inter-class topology in continual test-time adaptation},
  author={Ni, Chenggong and Lyu, Fan and Tan, Jiayao and Hu, Fuyuan and Yao, Rui and Zhou, Tao},
  booktitle={Proceedings of the Computer Vision and Pattern Recognition Conference},
  pages={15319--15328},
  year={2025}
}

@article{hendrycks2019benchmarking,
  title={Benchmarking neural network robustness to common corruptions and perturbations},
  author={Hendrycks, Dan and Dietterich, Thomas},
  journal={arXiv preprint arXiv:1903.12261},
  year={2019}
}

@article{lei2023comprehensive,
  title={A comprehensive survey of dataset distillation},
  author={Lei, Shiye and Tao, Dacheng},
  journal={IEEE Transactions on Pattern Analysis and Machine Intelligence},
  volume={46},
  number={1},
  pages={17--32},
  year={2023},
  publisher={IEEE}
}

@article{chakrabarty2023santa,
  title={Santa: Source anchoring network and target alignment for continual test time adaptation},
  author={Chakrabarty, Goirik and Sreenivas, Manogna and Biswas, Soma},
  journal={Transactions on Machine Learning Research},
  year={2023}
}

@inproceedings{boudiaf2022parameter,
  title={Parameter-free online test-time adaptation},
  author={Boudiaf, Malik and Mueller, Romain and Ben Ayed, Ismail and Bertinetto, Luca},
  booktitle={Proceedings of the IEEE/CVF Conference on Computer Vision and Pattern Recognition},
  pages={8344--8353},
  year={2022}
}

@article{adachi2024test,
  title={Test-time adaptation for regression by subspace alignment},
  author={Adachi, Kazuki and Yamaguchi, Shin'ya and Kumagai, Atsutoshi and Hamagami, Tomoki},
  journal={arXiv preprint arXiv:2410.03263},
  year={2024}
}

@inproceedings{choi2022improving,
  title={Improving test-time adaptation via shift-agnostic weight regularization and nearest source prototypes},
  author={Choi, Sungha and Yang, Seunghan and Choi, Seokeon and Yun, Sungrack},
  booktitle={European Conference on Computer Vision},
  pages={440--458},
  year={2022},
  organization={Springer}
}

@article{zhang2017mixup,
  title={mixup: Beyond empirical risk minimization},
  author={Zhang, Hongyi and Cisse, Moustapha and Dauphin, Yann N and Lopez-Paz, David},
  journal={arXiv preprint arXiv:1710.09412},
  year={2017}
}

@inproceedings{su2024unraveling,
  title={Unraveling batch normalization for realistic test-time adaptation},
  author={Su, Zixian and Guo, Jingwei and Yao, Kai and Yang, Xi and Wang, Qiufeng and Huang, Kaizhu},
  booktitle={Proceedings of the AAAI Conference on Artificial Intelligence},
  volume={38},
  number={13},
  pages={15136--15144},
  year={2024}
}

@inproceedings{xie2017aggregated,
  title={Aggregated residual transformations for deep neural networks},
  author={Xie, Saining and Girshick, Ross and Doll{\'a}r, Piotr and Tu, Zhuowen and He, Kaiming},
  booktitle={Proceedings of the IEEE conference on computer vision and pattern recognition},
  pages={1492--1500},
  year={2017}
}

@inproceedings{he2016deep,
  title={Deep residual learning for image recognition},
  author={He, Kaiming and Zhang, Xiangyu and Ren, Shaoqing and Sun, Jian},
  booktitle={Proceedings of the IEEE conference on computer vision and pattern recognition},
  pages={770--778},
  year={2016}
}

@inproceedings{lee2022dataset,
  title={Dataset condensation with contrastive signals},
  author={Lee, Saehyung and Chun, Sanghyuk and Jung, Sangwon and Yun, Sangdoo and Yoon, Sungroh},
  booktitle={International Conference on Machine Learning},
  pages={12352--12364},
  year={2022},
  organization={PMLR}
}

@inproceedings{du2023minimizing,
  title={Minimizing the accumulated trajectory error to improve dataset distillation},
  author={Du, Jiawei and Jiang, Yidi and Tan, Vincent YF and Zhou, Joey Tianyi and Li, Haizhou},
  booktitle={Proceedings of the IEEE/CVF conference on computer vision and pattern recognition},
  pages={3749--3758},
  year={2023}
}

@article{du2023sequential,
  title={Sequential subset matching for dataset distillation},
  author={Du, Jiawei and Shi, Qin and Zhou, Joey Tianyi},
  journal={Advances in Neural Information Processing Systems},
  volume={36},
  pages={67487--67504},
  year={2023}
}

@article{guo2023towards,
  title={Towards lossless dataset distillation via difficulty-aligned trajectory matching},
  author={Guo, Ziyao and Wang, Kai and Cazenavette, George and Li, Hui and Zhang, Kaipeng and You, Yang},
  journal={arXiv preprint arXiv:2310.05773},
  year={2023}
}

@inproceedings{liu2024dataset,
  title={Dataset distillation by automatic training trajectories},
  author={Liu, Dai and Gu, Jindong and Cao, Hu and Trinitis, Carsten and Schulz, Martin},
  booktitle={European Conference on Computer Vision},
  pages={334--351},
  year={2024},
  organization={Springer}
}

@article{lee2024selmatch,
  title={Selmatch: Effectively scaling up dataset distillation via selection-based initialization and partial updates by trajectory matching},
  author={Lee, Yongmin and Chung, Hye Won},
  journal={arXiv preprint arXiv:2406.18561},
  year={2024}
}

@inproceedings{yang2024neural,
  title={Neural spectral decomposition for dataset distillation},
  author={Yang, Shaolei and Cheng, Shen and Hong, Mingbo and Fan, Haoqiang and Wei, Xing and Liu, Shuaicheng},
  booktitle={European Conference on Computer Vision},
  pages={275--290},
  year={2024},
  organization={Springer}
}

@inproceedings{zhong2025towards,
  title={Towards stable and storage-efficient dataset distillation: Matching convexified trajectory},
  author={Zhong, Wenliang and Tang, Haoyu and Zheng, Qinghai and Xu, Mingzhu and Hu, Yupeng and Guan, Weili},
  booktitle={Proceedings of the Computer Vision and Pattern Recognition Conference},
  pages={25581--25589},
  year={2025}
}

@inproceedings{sajedi2023datadam,
  title={Datadam: Efficient dataset distillation with attention matching},
  author={Sajedi, Ahmad and Khaki, Samir and Amjadian, Ehsan and Liu, Lucy Z and Lawryshyn, Yuri A and Plataniotis, Konstantinos N},
  booktitle={Proceedings of the IEEE/CVF International Conference on Computer Vision},
  pages={17097--17107},
  year={2023}
}

@inproceedings{deng2024exploiting,
  title={Exploiting inter-sample and inter-feature relations in dataset distillation},
  author={Deng, Wenxiao and Li, Wenbin and Ding, Tianyu and Wang, Lei and Zhang, Hongguang and Huang, Kuihua and Huo, Jing and Gao, Yang},
  booktitle={Proceedings of the IEEE/CVF Conference on Computer Vision and Pattern Recognition},
  pages={17057--17066},
  year={2024}
}

@article{li2025hyperbolic,
  title={Hyperbolic dataset distillation},
  author={Li, Wenyuan and Li, Guang and Maeda, Keisuke and Ogawa, Takahiro and Haseyama, Miki},
  journal={arXiv preprint arXiv:2505.24623},
  year={2025}
}

@inproceedings{cui2025optical,
  title={Optical: Leveraging optimal transport for contribution allocation in dataset distillation},
  author={Cui, Xiao and Qin, Yulei and Zhou, Wengang and Li, Hongsheng and Li, Houqiang},
  booktitle={Proceedings of the Computer Vision and Pattern Recognition Conference},
  pages={15245--15254},
  year={2025}
}

@inproceedings{li2025diversity,
  title={Diversity-Enhanced Distribution Alignment for Dataset Distillation},
  author={Li, Hongcheng and Zhou, Yucan and Gu, Xiaoyan and Li, Bo and Wang, Weiping},
  booktitle={Proceedings of the IEEE/CVF International Conference on Computer Vision},
  pages={3747--3756},
  year={2025}
}

@inproceedings{zhu2024reshaping,
  title={Reshaping the online data buffering and organizing mechanism for continual test-time adaptation},
  author={Zhu, Zhilin and Hong, Xiaopeng and Ma, Zhiheng and Zhuang, Weijun and Ma, Yaohui and Dai, Yong and Wang, Yaowei},
  booktitle={European Conference on Computer Vision},
  pages={415--433},
  year={2024},
  organization={Springer}
}

@inproceedings{liu2024continual,
  title={Continual-mae: Adaptive distribution masked autoencoders for continual test-time adaptation},
  author={Liu, Jiaming and Xu, Ran and Yang, Senqiao and Zhang, Renrui and Zhang, Qizhe and Chen, Zehui and Guo, Yandong and Zhang, Shanghang},
  booktitle={Proceedings of the IEEE/CVF Conference on Computer Vision and Pattern Recognition},
  pages={28653--28663},
  year={2024}
}

@inproceedings{yang2024versatile,
  title={A versatile framework for continual test-time domain adaptation: Balancing discriminability and generalizability},
  author={Yang, Xu and Chen, Xuan and Li, Moqi and Wei, Kun and Deng, Cheng},
  booktitle={Proceedings of the IEEE/CVF Conference on Computer Vision and Pattern Recognition},
  pages={23731--23740},
  year={2024}
}

@article{krizhevsky2009learning,
  title={Learning multiple layers of features from tiny images},
  author={Krizhevsky, Alex and Hinton, Geoffrey and others},
  year={2009},
  publisher={Toronto, ON, Canada}
}

@inproceedings{deng2009imagenet,
  title={Imagenet: A large-scale hierarchical image database},
  author={Deng, Jia and Dong, Wei and Socher, Richard and Li, Li-Jia and Li, Kai and Fei-Fei, Li},
  booktitle={2009 IEEE conference on computer vision and pattern recognition},
  pages={248--255},
  year={2009},
  organization={Ieee}
}

@article{cazenavette2025dataset,
  title={Dataset Distillation for Pre-Trained Self-Supervised Vision Models},
  author={Cazenavette, George and Torralba, Antonio and Sitzmann, Vincent},
  journal={arXiv preprint arXiv:2511.16674},
  year={2025}
}

@inproceedings{shin2023loss,
  title = {Loss-curvature matching for dataset selection and condensation},
  author = {Shin, Seungjae and Bae, Heesun and Shin, Donghyeok and Joo,
            Weonyoung and Moon, Il-Chul},
  booktitle = {International Conference on Artificial Intelligence and
               Statistics},
  pages = {8606--8628},
  year = {2023},
  organization = {PMLR},
}

@article{nguyen2020dataset,
  title = {Dataset meta-learning from kernel ridge-regression},
  author = {Nguyen, Timothy and Chen, Zhourong and Lee, Jaehoon},
  journal = {arXiv preprint arXiv:2011.00050},
  year = {2020},
}

@article{zhou2022dataset,
  title = {Dataset distillation using neural feature regression},
  author = {Zhou, Yongchao and Nezhadarya, Ehsan and Ba, Jimmy},
  journal = {Advances in Neural Information Processing Systems},
  volume = {35},
  pages = {9813--9827},
  year = {2022},
}

@inproceedings{jeong2026multimodal,
  title={Multimodal Distribution Matching for Vision-Language Dataset Distillation},
  author={Jeong, Jongoh and Kwon, Hoyong and Kim, Minseok and Yoon, Kuk-Jin},
  booktitle={Proceedings of the IEEE/CVF Conference on Computer Vision and Pattern Recognition},
  pages={23072--23082},
  year={2026}
}

@article{jang2024talos,
  title={TALoS: Enhancing semantic scene completion via test-time adaptation on the line of sight},
  author={Jang, Hyun-Kurl and Kim, Jihun and Kweon, Hyeokjun and Yoon, Kuk-Jin},
  journal={Advances in Neural Information Processing Systems},
  volume={37},
  pages={74211--74232},
  year={2024}
}

@inproceedings{kim2025dc,
  title={DC-TTA: Divide-and-Conquer Framework for Test-Time Adaptation of Interactive Segmentation},
  author={Kim, Jihun and Kwon, Hoyong and Kweon, Hyeokjun and Jeong, Wooseong and Yoon, Kuk-Jin},
  booktitle={Proceedings of the IEEE/CVF International Conference on Computer Vision},
  pages={23279--23289},
  year={2025}
}

@inproceedings{kim2026bootstrapping,
  title={Bootstrapping Video Semantic Segmentation Model via Distillation-assisted Test-Time Adaptation},
  author={Kim, Jihun and Kwon, Hoyong and Kweon, Hyeokjun and Yoon, Kuk-Jin},
  booktitle={Proceedings of the IEEE/CVF Conference on Computer Vision and Pattern Recognition},
  pages={10766--10777},
  year={2026}
}

@inproceedings{kim2026test,
  title={Test-Time Training for LiDAR Semantic Segmentation under Corruption via Geometric Inlier Discrimination},
  author={Kim, Hyeonseong and Jang, Hyun-Kurl and Yoon, Kuk-Jin},
  booktitle={Proceedings of the IEEE/CVF Conference on Computer Vision and Pattern Recognition},
  pages={24206--24216},
  year={2026}
}

@inproceedings{lim2026and,
  title={When and Where to Reset Matters for Long-Term Test-Time Adaptation},
  author={Lim, Taejun and Hwang, Joong-Won and Lee, Kibok},
  booktitle={ICLR},
  year={2026}
}
